# A Comparative Experiment of Several Shape Methods in Recognizing Plants

A. Kadir[1], L.E. Nugroho[2], A. Susanto[3] and P.I. Santosa[4],

Department of Electrical Engineering, Gadjah Mada University
Yogyakarta, Indonesia
[1]`akadir@mti.ugm.ac.id`, [2]`lukito@mti.ugm.ac.id`,
[3]`adhisusanto@jmn.net.id`, [4]`insap@mti.ugm.ac.id`

## ABSTRACT

*Shape is an important aspects in recognizing plants. Several approaches have been introduced to identify objects, including plants. Combination of geometric features such as aspect ratio, compactness, and dispersion, or moments such as moment invariants were usually used toidentify plants. In this research, a comparative experiment of 4 methods to identify plants using shape features was accomplished. Two approaches have never been used in plants identification yet, Zernike moments and Polar Fourier Transform (PFT), were incorporated. The experimental comparison was done on 52 kinds of plants with various shapes. The result, PFT gave best performance with 64% in accuracy and outperformed the other methods.*

## 1. INTRODUCTION

In pattern recognition and images retrieval, shape is one of the important aspect used to characterize objects, beside of colours and textures. Actually, various approaches have been incorporated in object recognition or images retrieval [1]. According to [2], past research in recognizing objects can be broadly classified into two categories: a) contour-based and b) region-based approaches. The disadvantage of the contour-based features is the difficulty on finding the correct curvature points. Based on the contour of leaf, features were extracted to differentiate species. However, contour of leaves have variations even in the same species [3]. Based on this situation, statistical properties is recommended. In practices, researches choose one or combination of methods to recognize objects. For example, Mercimek, et al. [4] used moment invariants to identify real objects. This moment was also used to recognize three kinds of aeroplanes [5]and to detect coconuts [6]. For plants identification purpose, Wu, et al. [7] used shape slimness, defined as ratio of length to width of leaves, shape roundness, defined as ratio of area of leaf image and perimeter of leaf contour, and shape solidity, defined as ratio of the internal area connecting to valley points and the external area connecting the top points. They also used moment invariants for additional features. Other research [8] was also used aspect ratio (shape slimness) and other basic geometric features to recognize plants.Lee and Chen [2] used aspect ratio, compactness, centroid, and vertical and horizontal projections.Meanwhile, Zulkifli [9] used invariant moments and General Regression Neural Network and worked on 10 kinds of leaves.

Zernike moments form part of the general theory of the geometrical moments. They were introduced initially by F. Zernike in 1934 [10]. These moments have been used in several applications such as face detection [10], fingerprint recognition [11], and character recognition [12]. According to [1], Zernike moments have 3 advantages, (1) rotation invariance, the magnitudes of Zernike moments are invariant to rotation, (2) robustness, they are robust to noise and minor variations in shape, and (3) expressiveness, they have minimum information redundancy since the basis is orthogonal. Meanwhile, PFT, introduced by D. Zhang in 2002, is still uncommon for images retrieval. However based on [13], PFT is better than Zernike moments. Therefore a study is necessary to be made to show experimental performance of such approaches, especially for certain objects.

In this paper, we propose to compare 4 methods in recognizing plants especially using shape features in preparation for further researches. Two of the methods, Zernike moment and Polar Fourier Transform, are never used in plants identification before. This paper organized as follows: Section 2 describes geometric features, moment invariants, Zernike moments, and PFT, Section 3 explains how the experiments were accomplished, Section 4 yields the experimental results, and Section 5 concludes the results.

## 2. FEATURES FOR SHAPE RECOGNTION

### 2.1. Geometric Features

Two geometric features commonly used in leaves recognition are slimness and roundness. Slimness (sometime called as aspect ratio) is defined as follow:

$$slimness = \frac{l_1}{l_2} \qquad (1)$$

where $l_1$ is the width of a leaf and $l_2$ is the length of a leaf (Fig. 1).

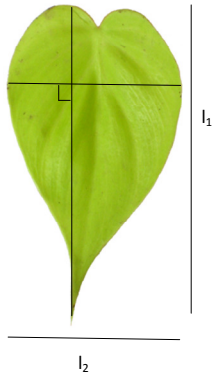

**Figure 1. Parameters for slimness of leaf**

Roundness (or compactness) is a feature defined as:

$$roundness = \frac{4\pi A}{P^2} \qquad (2)$$

where A is the area of leaf image and P is the perimeter of leaf contour.

Dispersion (irregularity) is another feature suggested by Nixon &Aguado [14] to deal with an object that has irregular shape such as the leaf in Fig. 2. This feature is defined as

$$IR(S) = \frac{\max(\sqrt{(x_i-\bar{x})^2+(y_i-\bar{y})^2})}{\min(\sqrt{(x_i-\bar{x})^2+(y_i-\bar{y})^2})} \qquad (3)$$

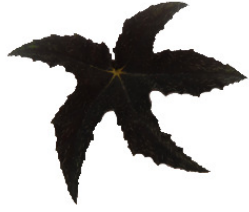

**Figure 2. Leaf with irregular shape**

The Eq. 3 defines the ratio between the radius of the maximum circle enclosing the region and the maximum circle that can be contained in the region. Therefore, the measure will increase as the region spreads. However, dispersion has a disadvantage. It is insensitive to slight discontinuity in the shape, such as crack in a leaf [14].

## *2.2 Moment Invariants*

Seven moments proposed by Hu [15] are very useful to capture shape of a leaf. Features based on this moments has been used in several experiments [4][7]. These moments are invariant under the actions of translation, scaling, and rotation. Computation is done as below.

$$\begin{aligned}
\emptyset_1 &= \eta_{20} + \eta_{02} \\
\emptyset_2 &= (\eta_{20} - \eta_{02})^2 + (2\eta_{02})^2 \\
\emptyset_3 &= (\eta_{30} - 3\eta_{12})^2 + (\eta_{03} - 3\eta_{21})^2 \\
\emptyset_4 &= (\eta_{30} + \eta_{12})^2 + (\eta_{03} + \eta_{21})^2 \\
\emptyset_5 &= (\eta_{30} - 3\eta_{12})(\eta_{30} + \eta_{12})\left[(\eta_{30} + \eta_{12})^2 - 3(\eta_{21} + \eta_{03})^2\right] + \\
&\quad (\eta_{03} - 3\eta_{12})(\eta_{03} + \eta_{21})\left[(\eta_{03} + \eta_{12})^2 - 3(\eta_{12} + \eta_{30})^2\right] \\
\emptyset_6 &= (\eta_{20} - \eta_{02})\left[(\eta_{30} + \eta_{12})^2 - (\eta_{21} + \eta_{03})^2\right] + \\
&\quad 4\eta_{11}(\eta_{30} + \eta_{12})(\eta_{03} + \eta_{21}) \\
\emptyset_7 &= (3\eta_{21} - \eta_{03})(\eta_{30} + \eta_{12})\left[(\eta_{30} + \eta_{12})^2 - 3(\eta_{21} + \eta_{03})^2\right] \\
&\quad (\eta_{30} - 3\eta_{12})(\eta_{21} + \eta_{03})\left[(\eta_{03} + \eta_{21})^2 - 3(\eta_{30} + \eta_{12})^2\right]
\end{aligned} \quad (4)$$

In this case,

$$\eta_{ij} = \frac{\mu_{pq}}{\mu^\gamma{}_{00}}, \gamma = \frac{i+j+2}{2} \quad (5)$$

$$\mu_{ij} = \sum_{x=1}^{M}\sum_{y=1}^{N}(x-\bar{x})^i (y-\bar{y})^j I(x,y) \quad (6)$$

$$\bar{x} = \frac{M_{10}}{M_{00}}, \bar{y} = \frac{M_{01}}{M_{00}} \quad (7)$$

$$M_{ij} = \sum_{x=1}^{M}\sum_{y=1}^{N} x^i y^j I(x,y) \quad (8)$$

where I(x, y) is the intensity of a pixel in the coordinate (x, y).

## 2.3 Zernike Moments

Zernike Moments (ZM) are orthogonal moments [1], derived from set of complex Zernike polynomials:

$$V_{n,m}(x,y) = V_{n,m}(r\cos\theta, r\sin\theta) = R_{n,m}(r).e^{jm\theta} \qquad (9)$$

where $R_{n,m}(r)$ is the orthogonal radial polynomial:

$$R_{n,m}(r) = \sum_{s=0}^{(n-|m|)/2}(-1)^s \frac{(n-s)!}{s!\left(\frac{n-2s+|m|}{2}\right)!\left(\frac{n-2s-|m|}{2}\right)!} r^{n-2s} \qquad (10)$$

n = 0, 1, 2, …; 0 $\leq |m| \leq$ n; and n - $|m|$ is even.

The Zernike moment of order n with repetition m of shape region f(x, y) is given by:

$$Z_{n,m} = \frac{n+1}{\pi}\sum_r \sum_\theta f(r\cos\theta, r\sin\theta).R_{n,m}(r).e^{jm\theta} \qquad (11)$$

In this case, r $\leq$ 1.

## 2.4 Polar Fourier Transform

There are 2 kinds of PFT proposed by D. Zhang. One of them is defined as follow [8]:

$$PF_2(\rho,\emptyset) = \sum_r \sum_i f(r,\theta_i)\exp\left[j2\pi(\frac{r}{R}\rho + \frac{2\pi}{T}\emptyset)\right] \qquad (12)$$

where

- $0 \leq r < R$ dan $\theta_i = i(2\pi/T)$ ($0 \leq I < T$); $0 \leq \rho < R$, $0 \leq \emptyset < T$,
- R is radial frequency resolution,
- T is angular frequency resolution.

How to compute PFT described as follow. For example, there is an image I = {f(x, y); 0$\leq$x<M, 0$\leq$y<N}. Firstly, the image is converted from Cartesian space to polar space $I_p$ = {f(r,$\theta$); 0$\leq$r<R, 0$\leq\theta < 2\pi$ }, where R is the maximum radius from centre of the shape. The origin of polar space becomes as centre of space to get translation invariant. The centroid ($x_c$, $y_c$) calculated by using formula:

$$x_c = \frac{1}{M}\sum_{x=0}^{N-1} x, y_c = \frac{1}{M}\sum_{y=0}^{M-1} y \qquad (13)$$

In this case, (r, ө ) is computed by using:

$$r = \sqrt{(x-x_c)^2 + (y-y_c)^2}, \theta = \arctan\frac{y-y_c}{x-x_c} \qquad (14)$$

Rotation invariance is achieved by ignoring the phase information in the coefficient. Consequently, only the magnitudes of coefficients are retained. Meanwhile, to get scale invariance, the first magnitude value is normalized by the area of the circle and all the magnitude values are normalized by the magnitude of the first coefficient. So, the shape descriptors are:

$$FD = \{\frac{PF(0,0)}{2\pi r^2}, \frac{PF(0,1)}{PF(0,0)}, \ldots, \frac{PF(0,n)}{PF(0,0)}, \ldots, \frac{PF(m,0)}{PF(0,0)}, \ldots, \frac{PF(m,n)}{PF(0,0)}\} \tag{15}$$

where m is the maximum number of the radial frequencies and m is the maximum number of angular frequencies.

## 3. MECHANISM OF LEAVES RETRIEVAL

Schema of plant identification is presented in Fig. 3. Features extracted from the leaf of query and each leaf in the database is used in calculating Euclidean distance to represent a rank. The Euclidean distance is computed using formula

$$d(Q,T) = \sqrt{\sum_{i=0}^{N-1}(Q_i - T_i)^2} \tag{16}$$

where N is the number of features, Q represents the features of query and T represents the features of leaf in the database. The leaf with the smallest rank is the most similar one.

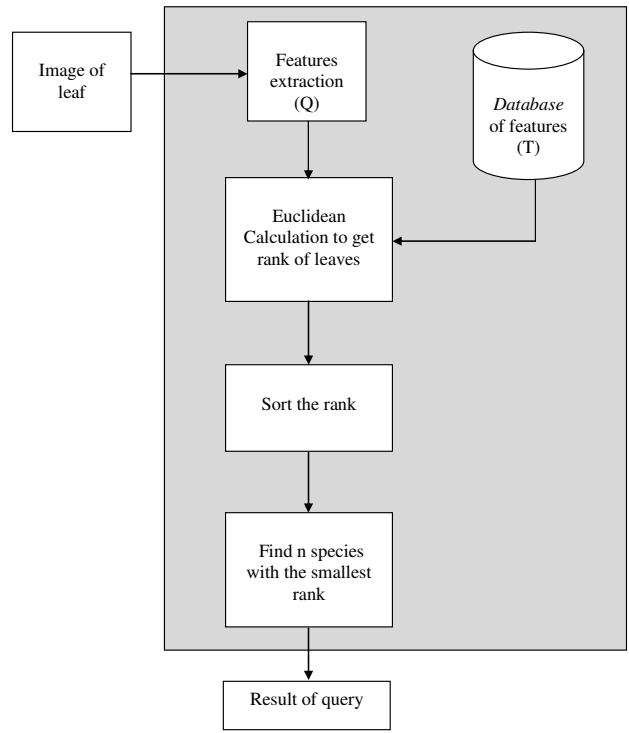

**Figure 3. Schema of plant identification**

In order to obtain performance of features, formula below was used:

$$Performance = \frac{Number\ of\ relevant\ images}{Total\ number\ of\ query} \tag{17}$$

Since several plants have similar shapes, there were three performance used in this experiments. First performance, called $p_1$, compared the testing leaf with the leaf of results with smallest rank. If the both of the leavescome fromthe same species then number of relevant images was increased by one. The second performance, called $p_3$, compared the testing leaf with third leaves of results with smallest rank. If the testing leaf is same species with one of three leaves then the number of relevant images was increased by one. The third performance, called $p_5$, compared the testing leaf with fifth leaves of results with smallest rank. If the testing leaf is same species with one of five leaves then the number of relevant images was increased by one. Performance $p_3$ and $p_5$ could be used as consideration when others aspects (colours and textures) will be included.

## 4. EXPERIMENTAL RESULTS

There were 50 kinds of plants with various colors and shapes used in this testing. All leaves came from our collection. Several of them have variegated leaves. Figure 3 shows the example of the leaves.

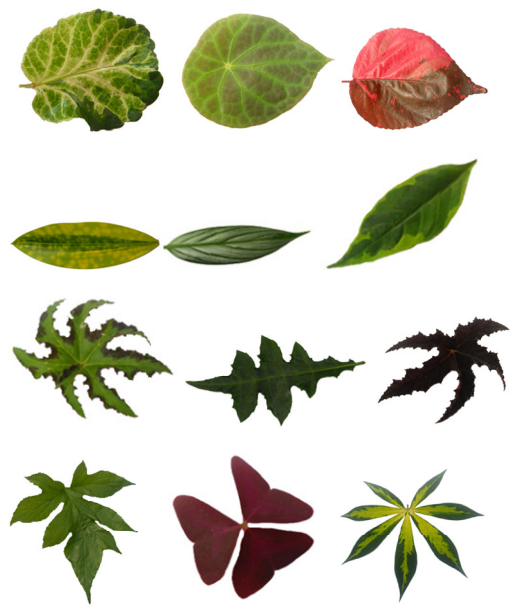

**Figure 4. Sampel of leaves**

Each plant in the database was represented by 20 samples. For testing purpose, 5 different leaves per plants were used.

In the experiment, we used Zernike moment with orde 7 and PFT with radial frequencies = 4 and angular frequencies = 6. Tabel 1 shows the result of several methods included in the experiments. Combination of geometric features, slimness, roundness, and dispersion did not give a good result. Zernike moments did not yield better solution than moment invariants. However, PFT gave a prospective result.

**Table 1. Performance of identification of shape of leaves**

| Method | Performance | | |
| --- | --- | --- | --- |
| | $P_1$ | $P_3$ | $P_5$ |
| Geometric Features | 10.80% | 24.40% | 36.40% |
| Moment invariants | 29.20% | 54.00% | 66.40% |

| Method | Performance | | |
|---|---|---|---|
| | $P_1$ | $P_3$ | $P_5$ |
| Moment invariant with normalization | 30.00% | 55.66% | 72.40% |
| Zernike moments | 18.80% | 40.40% | 51.60% |
| PFT | 64.00% | 86.40% | 93.20% |
| PFT+Moment invariants | 62.00% | 84.00% | 89.60% |

The last row in the table shows that combination of PFT and moment invariant did not improve the performance.

## 5. CONCLUSIONS

Based on the performance of several methods tried in this experiments, PFT outperforms among the others. This result shows that PFT has a good chance to be included in plants recognition. However, using shape only for plants identification is not enough. Therefore, for further researches, other aspects, such as colours and textures, should be incorporated to increase the performance of identification system, especially for foliage plants, where colours and patterns of leaf could not be ignored.Results on $p_3$ and $p_5$ gave an implicit sign for this action. Besides, based on visual observation of the query results, several plants that have similar shape but different colours were interchanged. Of course, by incorporating colours (and texture), the performance of system recognition can be improved.

**Authors**

**Abdul Kadir** received B.Sc in Electrical Engineering from Gadjah Mada University in 1987, M. Eng in Electrical Engineering from Gadjah Mada in 1998, Master of Management from Gadjah Mada University in 2004. His research interests include image processing, pattern recognition, and web-based applications.

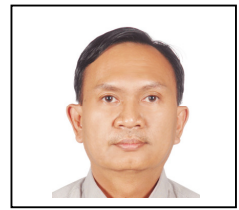

**Lukito Edi Nugroho** received B.Sc in Electrical Engineering from Gadjah Mada University in 1989, M.Sc. from James Cook University of North Queensland in 1994, Ph.D from School of Computer Science and Software Engineering, Monash University, in 2002.Hisresearch interests are software engineering, information systems, and multimedia.

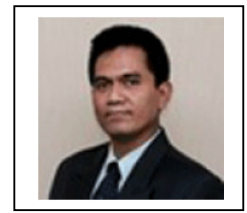

**Adhi Susanto** is a professor emeritus at Gadjah Mada University, Indonesia. He received Bachelor in Physics in 1964 from Gadjah Mada University, Master in Electrical Engineering in 1966 from University of California, Davis, USA, and Doctor of Philosophy in 1986 from University of California, Davis, USA. His research interests areas are electronics engineering, signal processing, and image processing.

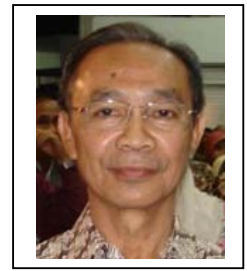

**Paulus Insap Santosa** obtained his undergraduate degree from Universitas Gadjah Mada in 1984,master degree from University of Colorado at Boulder in 1991, and doctorate degree from National University of Singapore in 2006. His research interests include human computer interaction and technology in Education.

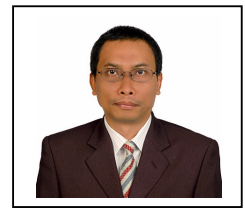